\documentclass[conference]{IEEEtran}
\IEEEoverridecommandlockouts
\usepackage{cite}
\usepackage{amsmath,amssymb,amsfonts}
\usepackage{algorithmic}
\usepackage{graphicx}
\usepackage{textcomp}
\usepackage{xcolor}

\newcommand\blfootnote[1]{%
  \begingroup
  \renewcommand\thefootnote{}\footnote{#1}%
  \addtocounter{footnote}{-1}%
  \endgroup
}

\def\BibTeX{{\rm B\kern-.05em{\sc i\kern-.025em b}\kern-.08em
    T\kern-.1667em\lower.7ex\hbox{E}\kern-.125emX}}
\begin{document}

\title{Dyna-DM: Dynamic Object-aware Self-supervised Monocular Depth Maps\\
}

\author{\IEEEauthorblockN{Kieran Saunders}
\IEEEauthorblockA{\textit{Department of Computer Science} \\
\textit{Aston University}\\
Birmingham, United Kingdom \\
190229315@aston.ac.uk}
\and
\IEEEauthorblockN{George Vogiatzis}
\IEEEauthorblockA{\textit{Department of Computer Science} \\
\textit{Loughborough University}\\
Leicestershire, United Kingdom \\
g.vogiatzis@lboro.ac.uk}
\and
\IEEEauthorblockN{Luis J. Manso}
\IEEEauthorblockA{\textit{Department of Computer Science} \\
\textit{Aston University}\\
Birmingham, United Kingdom \\
l.manso@aston.ac.uk}
}

\maketitle

\begin{abstract}
Self-supervised monocular depth estimation has been a subject of intense study in recent years, because of its applications in robotics and autonomous driving. Much of the recent work focuses on improving depth estimation by increasing architecture complexity.
This paper shows that state-of-the-art performance can also be achieved by improving the learning process rather than increasing model complexity.
More specifically, we propose
(i) disregarding \textit{small} potentially dynamic objects when training,
and (ii) employing an appearance-based approach to separately estimate object pose for truly dynamic objects. We demonstrate that these simplifications reduce GPU memory usage by 29\% and result in qualitatively and quantitatively improved depth maps. 
The code is available at https://github.com/kieran514/Dyna-DM.
\end{abstract}


\IEEEpeerreviewmaketitle 

\section{Introduction}

The problem of depth estimation from visual data has recently received increased interest because of its emerging application in autonomous vehicles.
Using cameras as a replacement for LiDAR and other distance sensors leads to cost efficiencies and can improve environmental perception. 

There have been several attempts at visual depth estimation, for example, using stereo images \cite{godard2017unsupervised, chang2018pyramid, li2019stereo} and optical flow \cite{ilg2017flownet, wang2018occlusion}. Other methods focus on supervised monocular depth estimation, which requires large amounts of ground truth depth data. Examples of supervised monocular architectures can be found in \cite{bhat2021adabins, song2021monocular, ranftl2021vision}. Ground truth depth maps such as those available in the KITTI dataset \cite{geiger2013vision} are usually acquired using LiDAR sensors. Supervised methods can be very precise but require expensive sensors and post-processing to obtain clear ground truth. This paper focuses on self-supervised monocular depth estimation, as it can provide the most affordable physical set-up and does not require annotated data.

\blfootnote{979-8-3503-0121-2/23/\$31.00 ©2023 IEEE}

Self-supervised methods exploit information acquired from motion. Examples of architectures using this method are given in \cite{godard2019digging, andraghetti2019enhancing, guizilini20203d, shu2020feature}. While supervised monocular depth estimation currently outperforms self-supervised methods, their performance is converging towards that of supervised ones.
Additionally, research has shown that self-supervised methods are better at generalizing across a variety of environments \cite{TENDLE2021100124} (\textit{e.g.}, indoor/outdoor, urban/rural scenes).


\begin{figure}[t]
\begin{center}
   \includegraphics[width=0.85\linewidth]{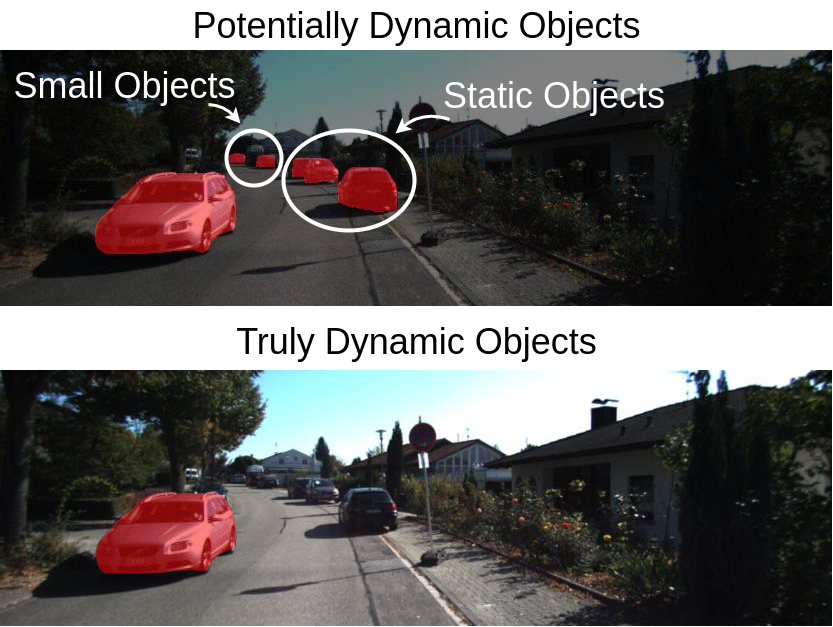}
\end{center}
   \caption{Dynamic objects have been consistently ignored in self-supervised monocular depth estimations. Our proposed method, Dyna-DM, isolates truly dynamic objects in a scene using an appearance-based approach.}
\label{fig:1}
\end{figure}

Many self-supervised monocular methods assume the entire 3D world is a rigid scene, thus ignoring objects that move independently. While convenient for computational reasons, in practice this assumption is frequently violated, leading to inexact pose and depth estimations. To account for each independently moving object, we would need to estimate the motion of the camera and each object separately. Unfortunately, slow-moving objects and deforming dynamic objects such as pedestrians make this a very challenging problem. 

Attempts to account for \emph{potentially dynamic objects} by detecting them and estimating their motion independently, which is the approach followed in this paper, can be found in \cite{casser2019depth} and \cite{lee2021learning}. However, not all potentially dynamic objects in a scene will be moving at any given time (\textit{e.g.}, cars parked on the side of the road). Estimating each potentially dynamic objects' motion is not only computationally wasteful, but more importantly, can cause significant errors in depth estimation (see Section~\ref{sec:results}). As these datasets predominantly contain static objects they cause the object motion network to be biased toward static objects.

Self-supervised models are generally trained on the KITTI \cite{geiger2013vision} and CityScape \cite{cordts2016cityscapes} datasets, where the latter is known to have a large amount of potentially dynamic objects.
Previous literature reports that truly dynamic objects are relatively infrequent (see Figure~\ref{fig:1}) and that static objects represent around 86\% of the pixels in KITTI test images~\cite{voicila2022instance}.
In addition to being computationally wasteful, estimating static objects' motion can reduce accuracy as seen in our experiments, Table \ref{table:8}.
To address this and other issues, the paper provides the following contributions:
\begin{itemize}
  \item A computationally efficient appearance-based approach to avoid estimating the motion of static objects.
  \item Experiments supporting the idea of removing small objects, as they tend to be far from the camera and their motion estimation does not bring significant benefits since they are ``almost rigid''.
\end{itemize}

\section{Related Works}

\subsection*{Self-Supervised Depth Estimation}
Self-supervised methods use image reconstruction between consecutive frames as a supervisory signal. Stereo methods use synchronized stereo image pairs to predict the disparity between pixels \cite{godard2017unsupervised}. For monocular video, the framework for simultaneous self-supervised learning of depth and ego-motion via maximising photometric consistency was introduced by Zhou \textit{et al.}~\cite{zhou2017unsupervised}. It uses ego-motion, depth estimation and an inverse warping function to find reconstructions for consecutive frames. This early model laid the foundation for self-supervised monocular depth estimation. Monodepth2 by Godard \textit{et al.}~\cite{godard2019digging} presented further improvements by masking stationary pixels that would cause infinite-depth computations. They cast the problem as the minimisation of image reprojection error while addressing occluded areas between two consecutive images. In addition, they employed multi-scale reconstruction, allowing for consistent depth estimations between the layers of the depth decoder. Monodepth2 is currently the most widely used baseline for depth estimation. 


\subsection*{Object Motion Estimation}
The previous methods viewed dynamic objects as a nuisance, masking non-static objects to find ego-motion. Later methods such as \cite{casser2019depth, lee2021learning} showed that there is a performance improvement when using a separate object-motion estimation for each potentially dynamic object. Casser \textit{et al.} \cite{casser2019depth} estimated object motion for potentially dynamic objects given predefined segmentation maps, cancelling the camera motion to isolate object motion estimations. Furthermore, Insta-DM by Lee \textit{et al.}~\cite{lee2021learning} improved previous models by using forward warping for dynamic objects and inverse warping for static objects to avoid stretching when warping dynamic objects. Additionally, this method uses a depth network (DepthNet) and pose network (PoseNet) based on a ResNet18 encoder-decoder structure following \cite{ranjan2019competitive}. Both methods treat all potentially dynamic objects in a scene as dynamic, therefore calculating object motion estimations for each individual object. 

There have been many appearance-based approaches using optical flow estimations \cite{lv2018learning,ranjan2019competitive}. Recent methods have combined geometric motion segmentation and appearance-based approaches \cite{yang2021learning}. These methods tackle the complex problem of detecting object motion, although their segmentation network is directly supervised by semantic segmentation data and motion estimations can be jittery and inaccurate.

Recently, Safadoust \textit{et al.} \cite{safadoust2021self} demonstrated detecting dynamic objects in a scene using an auto-encoder. While this seems promising for reducing the computation and improving depth estimation, it tends to struggle with texture-less regions such as the sky and plain walls. More complex methods have approached the detection of dynamic objects using transformers with a semantic and instance segmentation head \cite{voicila2022instance}, resulting in state-of-the-art performance in detecting dynamic objects but requiring heavy computation.

\begin{figure*}
\begin{center}
  \includegraphics[width=0.9\linewidth]{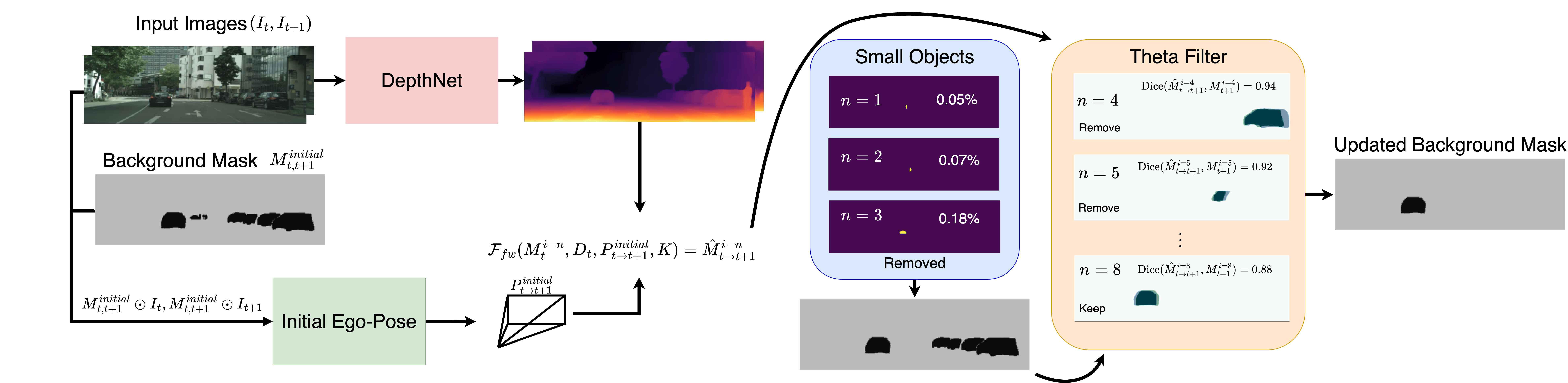}
\end{center}
  \caption{The input image and the background mask are used to calculate the initial ego-motion. Before removing static objects we remove small objects (objects less than 0.75\% of the image's pixel count) leading to an updated background mask. This mask is further processed using the theta filter, removing objects with a Dice value greater than $0.9$. This results in an updated background mask with truly dynamic objects only. }
\label{Arch}
\end{figure*}

\section{Method}
We propose Dyna-DM, an end-to-end self-supervised framework that learns depth and motion using sequential monocular video data. It introduces the detection of \emph{truly dynamic objects} using the Sørensen–Dice coefficient (Dice). 
We later discuss the use of object motion estimations for objects at a large distance from the camera.

\begin{figure*}
\begin{center}
  \includegraphics[width=0.925\linewidth]{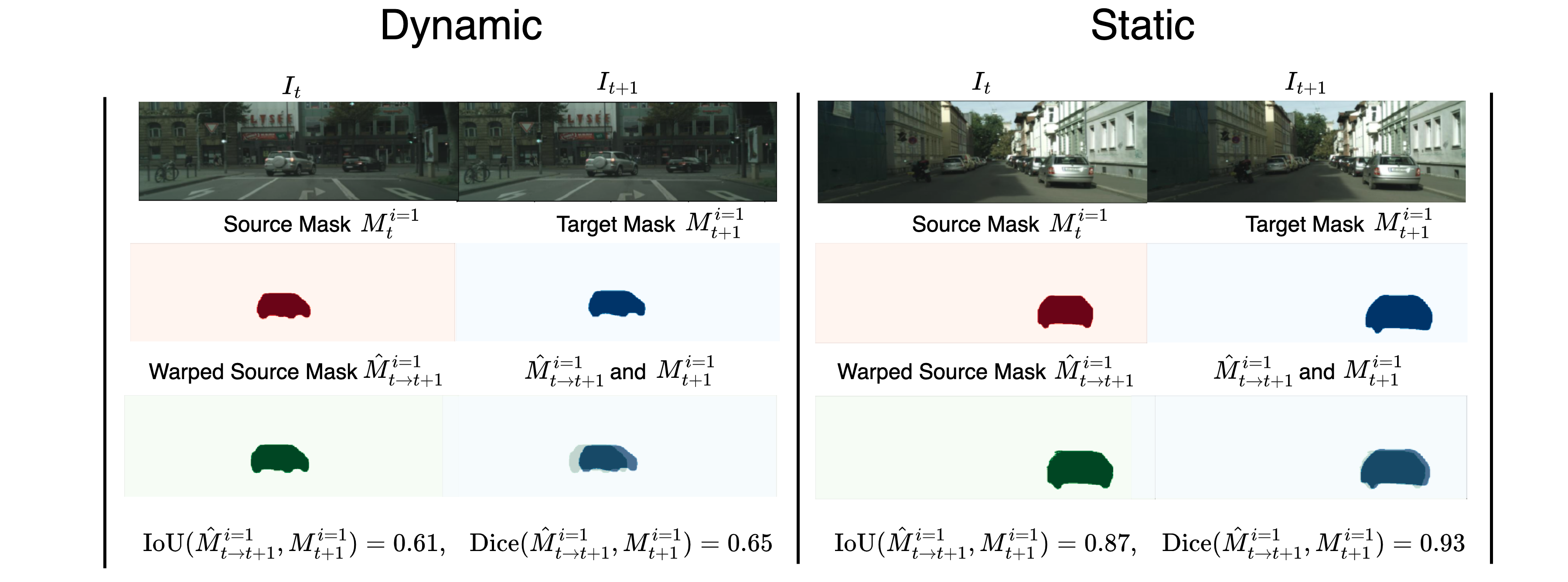}
\end{center}
  \caption{On the left, a dynamic object (in red) is warped using initial ego-motion, resulting in a warped source mask (green). Calculating the IoU and Dice coefficients between the target mask (blue) and warped source mask (green) a relatively small Dice value is observed. On the right, larger values for IoU and Dice were obtained following the same process.}
\label{fig:3}
\end{figure*}

\subsection{Self-supervised monocular depth estimation}\label{tradition}
To estimate depth in a self-supervised manner we consider consecutive RGB frames $(I_t, I_{t+1})$ in a video sequence. We then estimate depth $(D_{t}, D_{t+1})$ for both of these images using the inverse of the Disparity Network's (DispNet) output. To estimate the motion in the scene, two trainable pose networks (PoseNet) with different weights are used, one optimized for ego-motion and another to handle object-motion. 
The initial background images are computed by removing all potentially dynamic objects' masks $(M_{t}^i, M_{t+1}^i)$, corresponding to $(I_t, I_{t+1})$ respectively.
These binary masks are obtained using an off-the-shelf instance segmentation model Mask R-CNN \cite{he2017mask}. The next step is to estimate the ego-motion $(P_{t \to {t+1}}^{i=0},P_{{t+1} \to t}^{i=0})$ using these two background images. Following \cite{lee2021learning}, let us define inverse and forward warping, as follows:
\begin{equation}\label{24}
    \mathcal{F}_{iw}(I_t, D_{t+1}, P_{{t+1} \to t}, K) \to \hat{I}^{iw}_{t \to {t+1}},
\end{equation}
and
\begin{equation}\label{24}
   \mathcal{F}_{fw}(I_t, D_{t}, P_{t \to {t+1}}, K) \to \hat{I}^{fw}_{t \to {t+1}}
\end{equation}
where $K$ is the camera matrix.
Furthermore, we forward warp the image and mask using ego-motion, which we previously calculated, resulting in a forward warped image $\hat{I}_{t \to {t+1}}^{fw}$ and a forward warped mask $\hat{M}_{t \to {t+1}}^{fw, i=n}$. We feed the pixel-wise product of the forward warped mask and image and the pixel-wise product of the target mask $M_{t+1}^{i=n}$ and target image $I_{t+1}$ into the object motion network. Resulting in an object motion estimation for each potentially dynamic object $(P_{t \to {t+1}}^{i=n}, P_{{t+1} \to t}^{i=n})$, where $n=1,2,3\dots$ represents each potentially dynamic object. 
Using these object motion estimates we can inverse warp each object to give $\hat{I}_{t \to {t+1}}^{fw\to iw, i=n}$. The inverse warped background image is represented as $\hat{I}_{t \to {t+1}}^{iw, i=0}$.
The final warped image is a combination of the background warped image and all of the warped potentially dynamic objects as follows: 
\begin{equation}\label{24}
    \hat{I}_{t \to {t+1}}= \hat{I}_{t \to {t+1}}^{iw, i=0}+\sum_{n\in(1,2,...)}\hat{I}_{t \to {t+1}}^{fw\to iw, i=n}.
\end{equation}

\subsection{Truly Dynamic Objects}
To detect truly dynamic objects we first define an \emph{initial ego-motion network} that uses pre-trained weights from Insta-DM's ego-motion network. The binary masks from Insta-DM will also be used as our initial binary mask that contains all potentially dynamic objects.    
\begin{equation}
    M_{t,{t+1}}^{initial} = (1 - \cup_{n\in (1,2,...)}M_{t}^{i=n}) \cap (1 - \cup_{n\in (1,2,...)}M_{t+1}^{i=n})
\end{equation}
Then, similarly to section \ref{tradition}, we find the background image using pixel-wise multiplication $(\odot)$ and determine the initial ego-motion estimation between frames $I_t$ and $I_{t+1}$.
\begin{equation}
    P_{t \to {t+1}}^{initial} = Ego(M_{t,{t+1}}^{initial} \odot I_t, M_{t,{t+1}}^{initial} \odot I_{t+1})
\end{equation}
We then forward warp all of these masked potentially dynamic objects using this initial ego-motion. 
\begin{equation}
    \hat{M}_{t \to {t+1}}^{i=n} = F_{fw}(M_{t}^{i=n},D_{t},P_{t \to {t+1}}^{initial},K)
\end{equation}

Assuming perfect precision for pose and depth implies that the object mask will have been warped to the object mask in the target image if the object is represented by ego-motion i.e. if the object is static. In other words, if an object is static, there will be a significant overlap between its warped mask (under initial ego-motion) and its mask on the target image. Conversely, if the object is truly dynamic, the warped and target masks will not match. This type of overlap or discrepancy can be captured using the Sørensen–Dice coefficient or the Jaccard Index, also known as Intersection over Union (IoU).
\begin{equation}
    \text{Dice}(\hat{M}_{t \to {t+1}}^{i=n}, M_{t+1}^{i=n}) = \frac{2|\hat{M}_{t \to {t+1}}^{i=n} \cap M_{t+1}^{i=n}|}{|\hat{M}_{t \to {t+1}}^{i=n}| + |M_{t+1}^{i=n}|} < \theta
\end{equation}
\begin{equation}
    \text{IoU}(\hat{M}_{t \to {t+1}}^{i=n}, M_{t+1}^{i=n}) = \frac{|\hat{M}_{t \to {t+1}}^{i=n} \cap M_{t+1}^{i=n}|}{|\hat{M}_{t \to {t+1}}^{i=n} \cup M_{t+1}^{i=n}|} < \theta
\end{equation}

Warping of dynamic and static objects using ego-motion is depicted in Figure~\ref{fig:3}.
Stationary objects have greater Dice values than dynamic objects.
Potentially dynamic objects that have Dice values lower than the selected value of theta $(\theta)$ will be classed as truly dynamic objects.
Testing both IoU and Dice, we found that the Dice coefficient led to more accurate dynamic object detection, therefore the proposed solution is to use the Dice coefficient.
We found optimal $\theta$ values around the range [0.8, 1].

Note however that the reasoning presented above is based on the assumption that depth and pose estimations are accurate. This is unrealistic in most circumstances so we can expect larger Dice values with dynamic objects. To mitigate this challenge we can use greater frame separation between the source and target frames. This is simply done by calculating the reconstruction between frames $t$ and $t+2$ rather than frames $t$ and $t+1$. This extra-temporal distance between the frames gives dynamic objects more time to diverge from where ego-motion will warp them to be, which is beneficial for slow-moving objects. Extra-temporal distance would result in smaller IoU and Dice values for dynamic objects. With modern camera frame rates, extra-temporal distance does not cause jitters, but leads to more consistency in depth and more exact pose. As these objects can be moving very slowly, calculating the reconstruction between larger frames can be beneficial in determining imperceptible object motion.

To remove static objects, the method decreasingly sorts all potentially dynamic objects based on their Dice values and selects the first 20 objects.
These objects are the most likely to be static and tend to be larger objects, for $\theta$ less than 1, these objects with large Dice values will be removed first. The discrepancy between these metrics at $\theta$ equal to 1 can be explained by the metrics choosing different objects as the first 20 objects.

In summary, we filter using Dice scores to remove all static objects in the initial binary mask, providing an updated binary mask that only contains truly dynamic objects. We refer to this filtering process as \emph{theta filtering}. This can be used to calculate an updated ego-motion estimation and object motion estimations for each truly dynamic object. 
\subsection{Small Objects}
We improve this method further by removing small objects.
Small objects are objects that take a small pixel count in an image. We define these objects as less than 1\% of an image's pixel count. Motion estimation for small objects tends to be either inaccurate or insignificant as these objects are frequently at a very large distance from the camera. The removal of small objects and static objects are depicted in Figure \ref{Arch}.
\subsection{Final Loss}
Photometric consistency loss $\mathcal{L}_{pe}$ will be our main loss function in self-supervision. Following Insta-DM \cite{lee2021learning}, we apply consistency to the photometric loss function using a weighted valid mask $V_{t\to {t+1}}(p)$ to handle invalid instance regions, view exiting and occlusion. 
\begin{multline}
 \mathcal{L}_{pe}=\sum_{p\in P}V_{t \to {t+1}}(p)\cdot[(1-\alpha) \cdot |I_{t+1}(p)-\hat{I}_{t \to {t+1}}(p)|_{1} \\ + \frac{\alpha}{2}(1-SSIM(I_{t+1}(p),\hat{I}_{t \to {t+1}}(p)))]
\end{multline}
The geometric consistency loss $\mathcal{L}_g$ is a combination of the depth inconsistency map $D_{t \to {t+1}}^{diff}$ and the valid instance mask $\hat{M}_{t \to {t+1}}$ as shown in \cite{lee2021learning};
\begin{equation}
    \mathcal{L}_g=\sum_{p\in P}\hat{M}_{t \to {t+1}}(p)\cdot D_{t \to {t+1}}^{diff}(p).
\end{equation}
To smooth textures and noise while holding sharp edges, we apply an edge-aware smoothness loss $\mathcal{L}_s$ proposed by \cite{ranjan2019competitive}.
\begin{equation}
    \mathcal{L}_s=\sum_{p\in P}(\nabla D_{t}(p)\cdot e^{-\nabla I_t(p)})^2
\end{equation}
Finally, as there will be trivial cases for infinite depth for moving objects that have the same motion as the camera as discussed in \cite{godard2019digging}, we use object height constraint loss $\mathcal{L}_h$ as proposed by \cite{casser2019depth} given by;
\begin{equation}
    \mathcal{L}_h=\sum_{n\in\{1,2,\dots\}}\frac{1}{\Bar{D}}\cdot |D\odot M^{i=n}-\frac{f_y\cdot p_h}{h^{i=n}}|_{1}
\end{equation}
Where $p_h$ is a learnable height prior, $h^{i=n}$ is the learnable pixel height of the $n^{th}$ object and $\Bar{D}$ is the mean estimate depth. 
The final objective function is a weighted combination of the previously mentioned loss functions:
\begin{equation}
    \mathcal{L}= \lambda_{pe}\mathcal{L}_{pe} + \lambda_g\mathcal{L}_g + \lambda_s\mathcal{L}_s + \lambda_h\mathcal{L}_h.
\end{equation}
We followed the path of reconstructing $I_{t+1}$ from $I_t$, although the reconstruction of $I_t$ from $I_{t+1}$ would follow the same process.

\section{Results}\label{sec:results}


\subsection{Experimental Setup}
\textbf{Testing:}
We will be using the KITTI benchmark following the Eigen et al. split \cite{eigen2014depth}. 
The memory usage and power consumption were recorded as the maximum memory used during training GPUm and maximum power used while training GPUp recorded using Weights \& Biases \cite{wandb}. For CityScape, the ground truth data is inaccurate, therefore we will record the loss in equation \eqref{28} to inform us of improvements when doing hyperparameter tuning.
\begin{equation}\label{28}
     \mathcal{L} = \lambda_{pe}\mathcal{L}_{pe} + \lambda_g\mathcal{L}_g + \lambda_s\mathcal{L}_s
\end{equation}
\textbf{Implementation Details:}
Pytorch \cite{paszke2017automatic} is used for all models, training the networks on a single NVIDIA RTX3060 GPU (requiring 7GB of VRAM) with the ADAM optimiser, setting $\beta_1=0.9$ and $\beta_2=0.999$. Input images have a resolution of 832$\times$256 and are augmented using random scaling, cropping and horizontal flipping. The mini-batch size is set to 1 and we will be training the network with 1000 randomly sampled batches in each epoch for 40 epochs. Initially setting the learning rate to $10^{-4}$, and using exponential learning rate decay with $gamma = 0.975$. Finally, the weights are set to $(\lambda_{pe}, \lambda_g, \lambda_s, \lambda_h)=(2, 1, 0.1, 0.02)$ with $\alpha = 0.85$ as defined in \cite{lee2021learning}.

Following from \cite{ranjan2019competitive} we use DispNet, which is an encoder-decoder. This auto-encoder uses single-scale training as \cite{bian2019unsupervised} suggests faster convergence and better results. 
Ego-PoseNet, Obj-PoseNet and Initial Ego-PoseNet are all based on multiple PoseNet auto-encoders, but they do not share weights.

\subsection{Tuning the Theta parameter}
This section focuses on selecting a value of the filtering parameter, $\theta$, used to determine if an object is dynamic. Theta filtering removes all objects that are determined as static, leaving only object motion estimations for dynamic objects. Using the IoU measure (Jaccard index) as our measure, we iterative select $\theta$ values as described in Table \ref{table:3}. This table demonstrates that a value of 0.9 leads to the greatest reduction in loss while also reducing memory and power usage. 
\begin{table}[h]
\caption{Using the Jaccard index to remove static objects with varying $\theta$ values.}
\begin{center}
\footnotesize
\begin{tabular}{| c | c | c | c |}
\hline
\textbf{$\theta$} & \multicolumn{3}{ c |}{\textbf{Jaccard Index}}  \\ 
\cline{2-4}
& Loss $\downarrow$ & GPUm $\downarrow$ & GPUp $\downarrow$ \\
\hline

1 & 1.220 & 9.08GB & 100W \\ \hline
0.95 & 1.217 & 8.90GB & 94.7W \\ \hline
0.9 & \textbf{1.216} & 8.54GB & 87.4W \\ \hline
0.85 & 1.233 & 8.29GB & \textbf{84.5W} \\ \hline
0.8 & 1.264 & \textbf{8.04GB} & 88.9W \\ \hline

\end{tabular}
\label{table:3}
\end{center}
\end{table}
Furthermore, we explore an increased intra-frame distance to handle slow-moving objects. In Table \ref{table:4}, again the optimal value of $\theta$ is shown to be 0.9. Now the loss is shown to be less than in Table \ref{table:3}, suggesting that extra-temporal distance leads to improvements in detecting dynamic objects.

\begin{table}[h]
\caption{Using the Jaccard index and extra-temporal distance to determine if an object is dynamic based on three frames rather than two frames.}
\begin{center}
\footnotesize
\begin{tabular}{| c | c | c | c |}
\hline
\textbf{$\theta$} & \multicolumn{3}{ c |}{\textbf{Jaccard Index + Extra-temporal Distance}}  \\ 
\cline{2-4}
& Loss $\downarrow$ & GPUm $\downarrow$ & GPUp $\downarrow$ \\
\hline

1 & 1.215 & 8.79GB & 99.6W \\ \hline
0.95 & 1.218 & 8.92GB & 97.5W \\ \hline
0.9 & \textbf{1.207} & 9.06GB & 97.4W \\ \hline
0.85 & 1.220 & 8.56GB & 96.9W \\ \hline
0.8 & 1.251 & \textbf{8.19GB} & \textbf{89.8W} \\ \hline

\end{tabular}
\label{table:4}
\end{center}
\end{table}
Replacing the Jaccard Index with the Sørensen–Dice coefficient, we get the optimal $\theta$ value at 0.9 of the evaluated $\theta$ values, as seen in Table \ref{table:5}. This measure gives an even greater reduction in loss, memory and power usage.
Tables \ref{table:3}, \ref{table:4}, \ref{table:5}, show loss increases when using $\theta$ values under $0.9$. Arguably, doing so increases the probability of misclassifying dynamic objects as static.
\begin{table}[h]
\caption{Using extra-temporal distance and Dice coefficient to detect dynamic objects.}
\begin{center}
\footnotesize
\begin{tabular}{| c | c | c | c |}
\hline
\textbf{$\theta$} & \multicolumn{3}{ c |}{\textbf{Dice coefficient + Extra-temporal Distance}}  \\ 
\cline{2-4}
& Loss $\downarrow$ & GPUm $\downarrow$ & GPUp $\downarrow$ \\
\hline

1 & 1.177 & 8.53GB & 97.758W \\ \hline
0.95 & 1.174 & 8.87GB & 94.977W \\ \hline
0.9 & \textbf{1.172} & 8.24GB & 92.802W \\ \hline
0.85 & 1.180 & 8.42GB & \textbf{82.233W} \\ \hline
0.8 & 1.189 & \textbf{8.05GB} & 85.354W  \\ \hline

\end{tabular}
\label{table:5}
\end{center}
\end{table}

Using these methods, we must determine how many potentially dynamic objects we will process for each scene in advance. Previously, Insta-DM \cite{lee2021learning} focused on a maximum of 3 potentially dynamic objects, whereas here we will use a maximum of 20.

\begin{table}[h!]
\caption{Training the KITTI dataset to analyse the optimal value of $\theta$ on the KITTI Eigen test set.}
\begin{center}
\footnotesize
\tabcolsep=0.08cm

\begin{tabular}{|c|c|c|c|c|c|c|}
\hline
\textbf{$\theta$} & \multicolumn{6}{ c |}{\textbf{Dice coefficient + Extra-temporal Distance (KITTI)}}  \\ 
\cline{2-7}
& AbsRel$\downarrow$ & SqRel$\downarrow$ & RMSE$\downarrow$ & $\delta < 1.25\uparrow$ & GPUm$\downarrow$ & GPUp $\downarrow$ \\
\hline

1 & 0.118&     0.775&4.975&     0.860& 7.54GB & 99.4W\\ \hline
0.95 & 0.120&     0.799&4.799&   0.862& 7.44GB & 98.8W\\ \hline
0.9 & 0.117&     0.786&\textbf{4.709}&     \textbf{0.868}& 7.37GB & 98.2W\\ \hline
0.85 & \textbf{0.116}&\textbf{0.748}&4.870&     0.861 & 6.94GB & \textbf{96.1W}\\ \hline
0.8 & 0.118&     0.770&4.816&     0.863 & \textbf{6.74GB} & 97.4W\\ \hline

\end{tabular}
\label{table:6}
\end{center}
\end{table}

We base our $\theta$ experimentation on the CityScape dataset as it is known to have more potentially dynamic objects than the KITTI dataset. This allows us to determine the optimal value of $\theta$ in a setting where this method is potentially more valuable. Now testing with the KITTI dataset we obtain Table \ref{table:6}. A $\theta$ value of 0.9 is again optimal in this dataset, also showing a reduction in power and memory usage as $\theta$ decreases. 
To explore more quantitative evidence, we took the first 500 potentially dynamic objects from the validation set and labelled them as either dynamic or non-dynamic. Then exporting the IoU and Dice coefficient values for each associated object. For these values, we iteratively modify $\theta$ and determine which value was optimal for detecting if the object was truly dynamic. Using a $\theta$ value of 0.9 with the Jaccard index the method was only accurate 65\% of the time. Whereas, when using the Dice coefficient, with a $\theta$ value of 0.9, the method was accurate 74\% of the time. 
Although $\theta=0.9$ seems to lead to the greatest loss reduction, the selection of this value is dependent on the user as smaller values may lead to more misclassifications but lead to greater GPU memory and power reductions, a trade-off between accuracy and GPU usage.  

\subsection{Small Objects}

\begin{table}[h!]
\caption{The removal of small objects in each image of size less than the first column's percentage value.}
\begin{center}
\footnotesize
\begin{tabular}{| c | c | c | c |}
\hline
\textbf{\%} & \multicolumn{3}{ c |}{\textbf{Removing Small Objects}}  \\ 
\cline{2-4}
& Loss $\downarrow$ & GPUm $\downarrow$ & GPUp $\downarrow$ \\
\hline

0 & 1.172 & 8.12GB & 94.345W \\ \hline
0.25 & 1.172 & 7.03GB & 88.437W \\ \hline
0.5 & 1.172 & 7.66GB & 85.442W \\ \hline
0.75 & \textbf{1.169} & \textbf{6.64GB} & \textbf{83.973W} \\ \hline
1 & 1.173 & 6.68GB & 88.189W \\ \hline

\end{tabular}
\label{table:7}
\end{center}
\end{table}

\begin{table*}[h]
\caption{Testing on the KITTI Eigen test set, trained on the Cityscape and KITTI datasets. }
\begin{center}
\small
\tabcolsep=0.11cm
\begin{tabular}{|l|c|c|c|c|c|c|c|c|c|}
\hline
\textbf{Methods} & \multicolumn{9}{ c |}{\textbf{Comparison Table}}  \\ 
\cline{2-10}
& AbsRel$\downarrow$ & SqRel$\downarrow$ & RMSE$\downarrow$ & RMSE log$\downarrow$ & $\delta < 1.25\uparrow$ & $\delta < 1.25^2\uparrow$ & $\delta < 1.25^3\uparrow$ & GPUm$\downarrow$ & GPUp $\downarrow$ \\
\hline

Monodepth2 \cite{godard2019digging} &  0.132 & 1.044 & 5.142 & 0.210 & 0.845 & 0.948 & 0.977 & 9GB & 112W\\ \hline
Struct2Depth \cite{casser2019depth} & 0.141 & 1.026 & 5.290 & 0.215 & 0.816 & 0.945 & 0.979 & 10GB & 116W\\ \hline
Insta-DM \cite{lee2021learning}  &0.121&     0.797&     4.779&     0.193&     0.858&     0.957&     \textbf{0.983}& 9.48GB & 107.9W\\ \hline
Dyna-DM (ours) & \textbf{0.115}&\textbf{0.785}& \textbf{4.698}& \textbf{0.192}& \textbf{0.871}&     \textbf{0.959}&     0.982 & \textbf{6.67GB} & \textbf{94.0W}\\ \hline

\multicolumn{10}{c}{$^{\mathrm{a}}$Dyna-DM uses a $\theta$ value $0.9$ with the Sørensen–Dice coefficient and removes all objects $\leq$ 0.75\% of the images pixel count.} \\

\multicolumn{10}{c}{$^{\mathrm{b}}$Monodepth2 and Struct2Depth in this case are trained just with the KITTI dataset.}

\end{tabular}
\label{table:8}
\end{center}
\end{table*}
\begin{table*}[h!]
\caption{Using the same setup, we train with just the CityScape dataset.}
\begin{center}
\small
\tabcolsep=0.11cm
\begin{tabular}{|l|c|c|c|c|c|c|c|c|c|}
\hline
\textbf{Methods} & \multicolumn{9}{ c |}{\textbf{CityScape Only}}  \\ 
\cline{2-10}
& AbsRel$\downarrow$ & SqRel$\downarrow$ & RMSE$\downarrow$ & RMSE log$\downarrow$ & $\delta < 1.25\uparrow$ & $\delta < 1.25^2\uparrow$ & $\delta < 1.25^3\uparrow$ & GPUm$\downarrow$ & GPUp $\downarrow$ \\
\hline

Insta-DM \cite{lee2021learning} &0.178&     1.312&     6.016&     0.257&     0.728&     0.916&     0.966& 10.58GB & 110.8W\\ \hline
Dyna-DM (ours) & \textbf{0.163}&     \textbf{1.259}&     \textbf{5.939}&     \textbf{0.244}&     \textbf{0.768}&     \textbf{0.926}&     \textbf{0.970}& \textbf{6.63GB} & \textbf{86.99W}\\ \hline

\end{tabular}
\label{table:9}
\end{center}
\end{table*}

Table \ref{table:7} demonstrates the removal of objects smaller than a specific percentage of the image's pixel count. We are using a $\theta$ value of 0.9 and the Sørensen–Dice coefficient with extra intra-frame distance. We observe that as the percent value increases we will be removing more objects and therefore reducing memory and power usage. But we see an increase in loss after 0.75\% as objects greater than this could represent small objects that are close to the camera, like children and animals. If they are small objects close to the camera then any object motion will cause significant increases in the loss as more pixels will be displaced compared to large objects far from the camera. This would be greatly detrimental if we ignored these objects, therefore, as suggested by the increasing loss, we will keep this value low to only account for objects at a far distance. 
We see that theta filtering with extra-temporal distance leads to the best improvements for all metrics. On the other hand, removing small objects leads to the best reduction in GPU usage while slightly improving accuracy.

\subsection{Comparison Table}
In summary, our method, which we refer to as Dyna-DM, removes all objects that occupy less than 0.75\% of the image's pixel count. Finally and most importantly, it determines which objects are dynamic using the Sørensen–Dice coefficient with a $\theta$ value of 0.9, therefore only calculating object motion estimations for these truly dynamic objects. We will be comparing our method with Monodepth2 \cite{godard2019digging}, Struct2Depth \cite{casser2019depth} and Insta-DM \cite{lee2021learning}. Dyna-DM has been initialised by weights provided by Insta-DM \cite{lee2021learning}. 
We train consecutively through CityScape and then the KITTI dataset for Insta-DM and Dyna-DM then test with the Eigen test split. Insta-DM and Struct2Depth will be using a maximum number of dynamic objects of 13 and Dyna-DM will use 20. As Dyna-DM filters out static and small objects, resulting in significantly less GPU usage, we can process more potentially dynamic objects compared to Insta-DM. The results are reported in Table \ref{table:8}.



Here we see significant improvements in all metrics except for $\delta < 1.25^3$ when comparing Dyna-DM to Insta-DM in Table \ref{table:8}. These metric improvements are accompanied by improvements in GPU usage. We most notably see a 29.6\% reduction in memory usage when training comparing Dyna-DM to Insta-DM, this allows us to improve the accuracy of our pose and depth estimations while requiring less computation.  


As we know that the CityScape dataset has more potentially dynamic objects than KITTI we can train Dyna-DM and Insta-DM on CityScape and test with the Eigen test split. Table \ref{table:9} demonstrates even greater improvements in all metrics when comparing Dyna-DM and Insta-DM. Our model can remove most static objects from these object motion estimations while isolating truly dynamic objects, thereby leading to significant improvements in pose estimation. This further improves the reconstructions and leads to improved depth estimations. 
When re-training Insta-DM using a maximum of 13 dynamic objects, we see worse results than reported in the original paper \cite{lee2021learning}. We believe this is because our method is able to interpret many potentially dynamic objects, whereas, Insta-DM results in worse object motion estimation for a larger maximum number of dynamic objects processed.


Note that in some real-life circumstances, we have the limitation that there will be unique dynamic objects like debris that would not be segmented in this case. Therefore for an efficient, truly autonomous vehicle, we would have to address these safety issues.

\section{Conclusion}
Dyna-DM reduces memory and power usage during training monocular depth-estimation while providing better accuracy in test time.
Although not all dynamic objects are always classified accordingly (e.g., slow-moving cars) Dyna-DM detects significant movements which are the ones which would cause the greatest reconstruction errors.
We believe that the best step forward is to make this approach completely self-supervised by detecting all dynamic objects, including debris, using an auto-encoder for safety and efficiency. This means we will have to be able to handle textureless regions and lighting issues which will be shown in future work.
Other further improvements include using depth maps to determine objects' distances, and removing object motion estimations for objects at larger distances.

\section*{Acknowledgment}
Most experiments were run on Aston EPS Machine Learning Server, funded by the EPSRC Core Equipment Fund, Grant EP/V036106/1.

\bibliographystyle{IEEEtranstyle}
\bibliography{mybibfile}


\end{document}